\documentclass[journal]{IEEEtran}
\usepackage{graphicx}
\usepackage{booktabs}    %导言区	
\usepackage{amsmath}
\usepackage{stfloats}
\usepackage{amssymb}
\usepackage{bbding}
\usepackage{notoccite}
\usepackage[numbers,sort&compress]{natbib}
\DeclareUnicodeCharacter{0301}{\'{e}}
\begin{document}
\title{A Multimodal Dangerous State Recognition and Early Warning System for Elderly with Intermittent Dementia}

\author{Liyun Deng,
        Lei Jin, % <-this % stops a space
        Guangcheng Wang,
        Quan Shi,
        Han Wang*% <-this % stops a space

\thanks{Liyun Deng and Lei Jin is an undergraduate student at the School of Transportation and Civil Engineering, Nantong University, located at 9 Sheyuan Road, Chongchuan District, Nantong. The email address is 	2133110153@stmail.ntu.edu.cn.}% <-this % stops a space
\thanks{Guangcheng Wang, Quan Shi and Han Wang are with Nantong University.Han Wang's email address is hanwang@ntu.edu.cn}% <-this % stops a space
}

\maketitle

\begin{abstract}
In response to the social issue of the increasing number of elderly vulnerable groups going missing due to the aggravating aging population in China, our team has developed a wearable anti-loss device and intelligent early warning system for elderly individuals with intermittent dementia using artificial intelligence and IoT technology. This system comprises an anti-loss smart helmet, a cloud computing module, and an intelligent early warning application on the caregiver's mobile device. The smart helmet integrates a miniature camera module, a GPS module, and a 5G communication module to collect first-person images and location information of the elderly. Data is transmitted remotely via 5G, FTP, and TCP protocols. In the cloud computing module, our team has proposed for the first time a multimodal dangerous state recognition network based on scene and location information to accurately assess the risk of elderly individuals going missing. Finally, the application software interface designed for the caregiver's mobile device implements multi-level early warnings. The system developed by our team requires no operation or response from the elderly, achieving fully automatic environmental perception, risk assessment, and proactive alarming. This overcomes the limitations of traditional monitoring devices, which require active operation and response, thus avoiding the issue of the digital divide for the elderly. It effectively prevents accidental loss and potential dangers for elderly individuals with dementia.
\end{abstract}


% \fi
\IEEEpeerreviewmaketitle

\section{Introduction}

\IEEEPARstart{B}{etween}  2015 and 2020, the number of missing people in China decreased, but the proportion of vulnerable groups, particularly the elderly and children, increased annually. According to the "White Paper on Missing Persons in China (2020)," the number of missing people in China reached 1 million in 2020, with about half of them being elderly. Approximately 72\% of these elderly individuals had memory impairments (25\% were diagnosed with dementia). Among the rescued missing elderly, 26\% went missing again, with 6\% going missing more than five times\cite{1}. Experts point out that in developed countries, dementia is the main cause of elderly individuals going missing. However, in China, the aging process is accompanied by large-scale population movements, resulting in a large number of elderly living alone. By 2020, the number of elderly living alone in China exceeded 118 million\cite{2}. The lack of care due to population mobility and the economic poverty of the elderly significantly increase the risk of going missing\cite{1}.

Meanwhile, memory decline is one of the most common issues among the elderly. As they age, their cognitive functions gradually deteriorate, making it easy for them to forget the way home. This study defines such elderly individuals as those with intermittent dementia. Caring for these elderly individuals is highly challenging. On one hand, they cannot be confined at home because numerous studies have shown that prolonged social isolation exacerbates negative emotions such as loneliness\cite{3,4,5} and anxiety\cite{6}, which are detrimental to their physical and mental health. On the other hand, they cannot be allowed to go out freely because caregivers cannot predict when they might experience an episode. Therefore, developing a wearable device and intelligent early warning system is crucial.This device can monitor the safety of elderly individuals in real-time when they go for walks, shopping, or other activities and send alerts to caregivers when necessary. In this way, caregivers can keep track of the elderly's safety status while also having time to attend to their own work and life, thereby reducing the burden on families and society.

Current methods for determining whether elderly individuals are in a dangerous state primarily rely on electronic fence settings or fall detection\cite{7,8,9,10}. Additionally, some novel methods include analyzing the historical movement patterns of the elderly to determine if their behavior is normal\cite{11}. However, these methods often rely on a single indicator when assessing the safety of the elderly, neglecting the impact of environmental factors on risk. They lack a comprehensive evaluation method that integrates information on the environment, time, and location to achieve a thorough risk assessment.

Currently, smart anti-loss devices for the elderly primarily use microcontrollers\cite{12,13}, Raspberry Pi, and similar core processors, combined with Global Positioning System (GPS), Beidou Navigation System\cite{14},Bluetooth\cite{15,16}, and Radio Frequency Identification(RFID) technology\cite{17,18} to achieve real-time positioning and data collection for the elderly. These devices transmit data to servers via Bluetooth, WiFi, and General Packet Radio Service (GPRS)\cite{19} wireless communication modules for safety analysis and integrate these technologies into wearable devices such as clothing, necklaces, or robots accompanying the elderly. Additionally, mobile apps have been developed for caregivers to view the elderly's location information and issue timely warnings in emergencies. 
However, these devices often lack stability and reliability when using Bluetooth or WiFi for long-distance communication outdoors. They also cannot intuitively display first-person perspective images of the elderly or accurately show the elderly's real-time location on a map in the mobile application.
\begin{figure*}[!t]
\centering
\includegraphics[width=\linewidth]{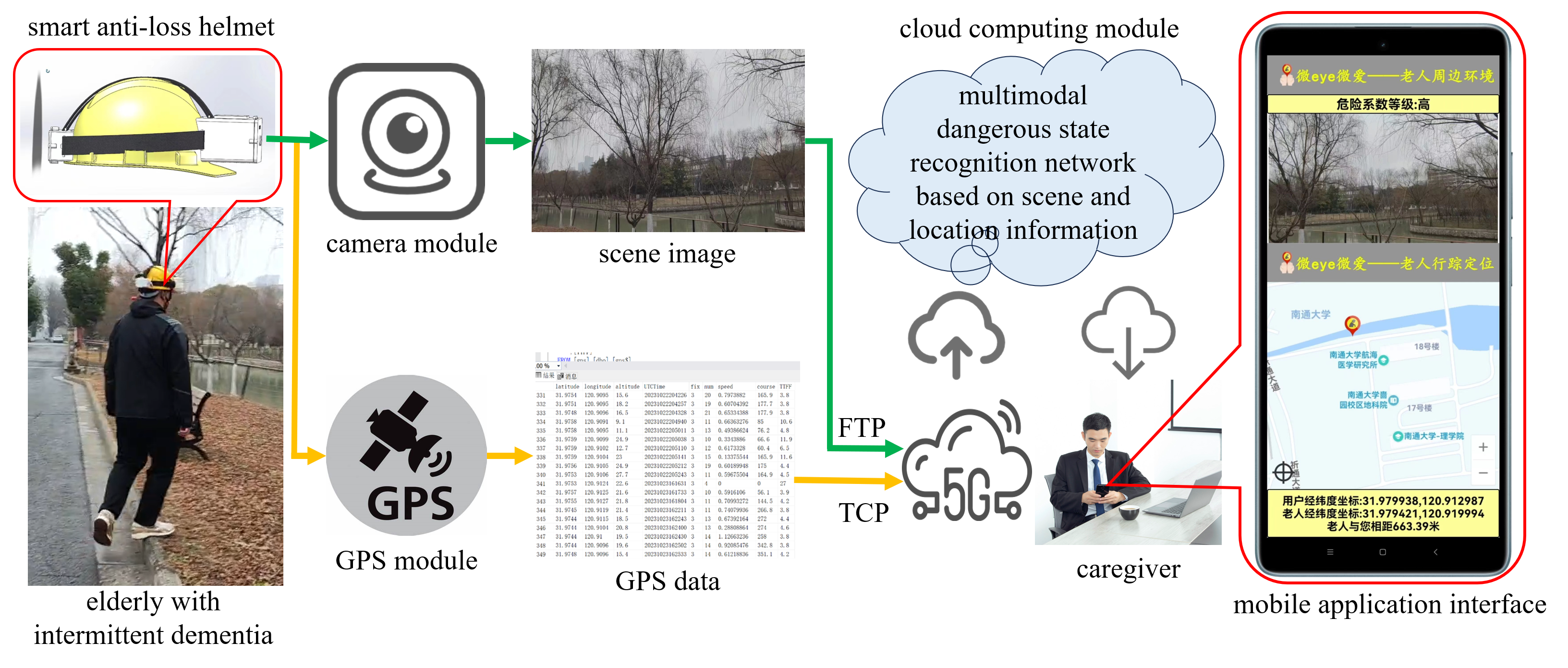}
\caption{Overall architecture and operation flowchart of the anti-loss and dangerous state recognition early warning system based on a smart helmet}
\label{fig1}
\end{figure*}
To address the limitations of existing methods and devices, this study proposes an innovative solution. We have designed a multimodal dangerous state recognition method and an early warning system. This system uses a smart anti-loss helmet integrated with an intelligent camera module to capture first-person perspective images and location information of the elderly in real-time. Utilizing 5G communication technology and FTP and TCP transmission protocols, the device efficiently transmits collected data to a cloud server. On the server side, we have built a multimodal dangerous state recognition network based on scene and location information, which comprehensively and real-time assesses the elderly's dangerous state by combining images and location data. Furthermore, we have developed a user-friendly and easy-to-operate anti-loss early warning software. This application can receive and display image data, location information, and dangerous state recognition results transmitted by the server in real-time. Based on the recognition results, the application can automatically trigger different levels of warnings without any operation from the elderly, ensuring timely and reliable safety alerts. The overall framework and operation process of this system are shown in Figure \ref{fig1}.

The main contributions of this paper include the following three points:
\subsubsection{Creation of a Small Dataset}
We created a small dataset containing scene images captured from the first-person perspective of the elderly and map images generated based on location information. Due to privacy concerns with the elderly's location data during their outdoor activities, there is currently a lack of publicly available elderly behavior datasets. Therefore, this study collected data in public places by simulating the outdoor behavior of elderly individuals.
\subsubsection{Proposal of a Multimodal Dangerous State Recognition Method}
This method includes the Map Generation Module (MGM), Unimodal Feature Extraction Module (UFE), Multimodal Feature Fusion Module (MFF), and Hazardous State Recognition Module (HSR). The design and implementation of this network allow for a more comprehensive identification and evaluation of the elderly's dangerous states compared to existing technologies.
\subsubsection{Design and Implementation of an Anti-Loss Dangerous State Recognition and Early Warning System Based on a Smart Helmet}
This system comprises three parts: a smart anti-loss helmet with real-time data collection capabilities, which protects the elderly's head while reliably transmitting data to the server; a multimodal dangerous state recognition network based on scene and location information, which analyzes and identifies potential dangerous states; and an Android-based intelligent early warning software that visually displays the elderly's perspective images and location information, and issues multi-level warnings based on the dangerous state recognition results.
 
\section{Related work}

In this section, we will review the development of current anti-loss smart devices and early warning systems, as well as the latest advancements in intelligent hazardous environment recognition technology.
\begin{figure*}[t]
\centering
\includegraphics[width=\linewidth]{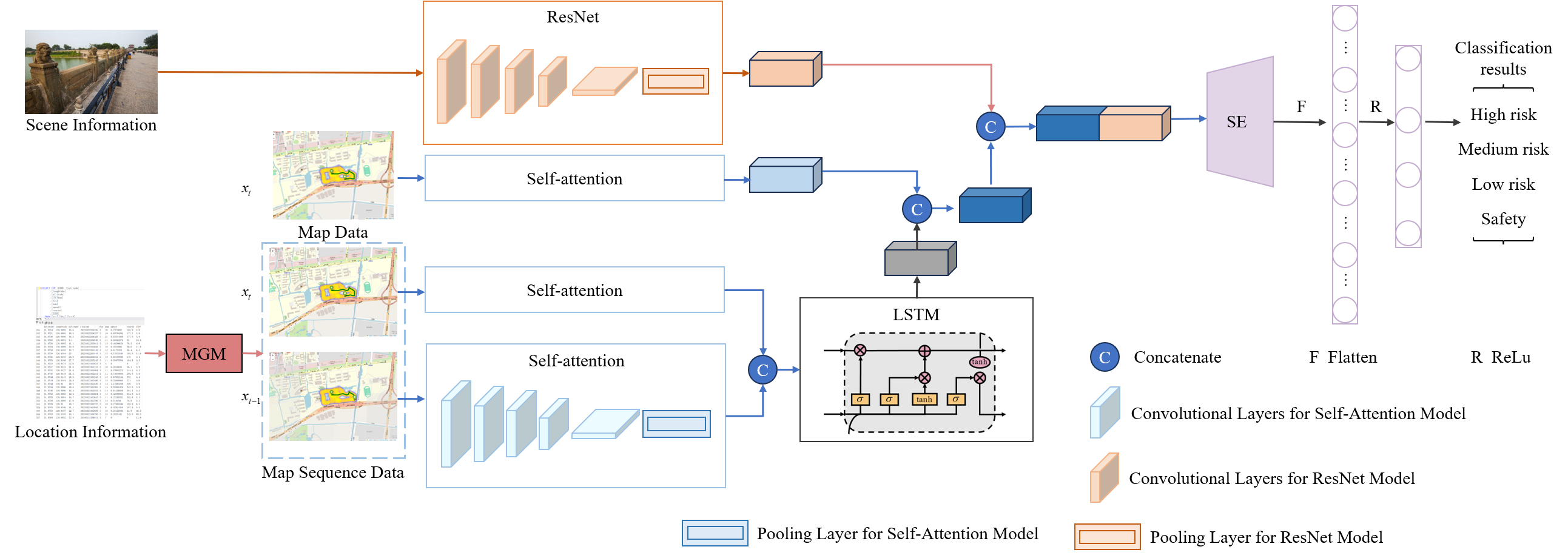}
\caption{The structure diagram of the scene-location multimodal dangerous state recognition network.}
\label{fig2}
\end{figure*}
\subsection{intelligent hazardous environment recognition methods}
Currently, in the field of hazardous environment recognition and evaluation, the primary models used include traditional models, machine learning models, and single-modal models in deep learning. However, the application of multimodal models in this field is relatively rare.

Traditional Models:
(1) Pixel Attribute Variation: Methods like tracking the color changes of pixels around a fire using OpenCV and NumPy, combining brightness and chroma changes to develop an Adaptive Target Detection Technology (ATDT) module that can detect fires in different environments by integrating video camera and sensor data to reduce false alarm rates\cite{20}.
(2) Multi-Criteria Decision Analysis (MCDA)\cite{21}: Methods like MCDA\cite{21} and techniques applying Best-Worst Method (BWM) and Multi-Attribute Decision Making (MARCOS) under Interval Type-2 Fuzzy Sets (IT2FSs) for hazardous area detection and risk identification\cite{22}.
(3) Weight Determination and Data Fusion: Techniques determining the weights of multiple data sources, integrating data features for dangerous state recognition, such as using the corresponding target feature dispersion factor of different sensor data\cite{23}, Analytic Hierarchy Process (AHP)\cite{24,25}, and I-AHP\cite{26}. 
(4) Mathematical Models:Directly identifying danger levels through mathematical models\cite{27,28,29,31,32,33}. Although these traditional methods have achieved some success in environmental risk assessment, they have significant limitations: they usually handle relatively simple mappings and struggle to deeply explore and utilize the deep semantic information in data.

Machine Learning Models:
Researchers often use simple neural networks, such as various Back Propagation (BP) neural networks\cite{34,35}, Bayesian networks\cite{36} , random forests\cite{37,38}, concept neural networks\cite{35}, and fully connected neural networks  for environmental and industrial risk assessment. Some researchers also combine statistical methods and decision tree models for risk assessment\cite{39}  or use machine learning classifiers to classify risk levels\cite{40}. While these machine learning models perform well in simple pattern recognition and classification problems, their algorithms' limitations may make them unsuitable for handling more complex data and problems.

Single-Modal Models:Many studies use single CNN, Yolov5\cite{41}, Elman neural networks\cite{42}, and graph neural networks\cite{18}  for hazard recognition. Researchers have proposed hybrid neural networks like CNN\_LSTM\cite{43}, instance segmentation\cite{44}, and object detection\cite{45}, analyzing the relative positions of people and hazardous objects to assess risk levels.

These studies demonstrate the diversity and efficiency of single-modal models in safety assessment and target detection. However, they also reveal that single-modal models can only process one type of data, limiting their potential in more complex or dynamic environments.

Multimodal Models:In multimodal model research, X. Xu et al. used affine transformation techniques and a fusion module (DMF) to combine infrared and RGB sensor data, aligning features and reducing differences between sensors for efficient target recognition, especially in detecting potential intrusions in railway safety, enhancing driving safety\cite{46}. Xing et al. developed a semantic segmentation model (FSA-UNet) based on UNet, combining remote sensing and street view image features to fully exploit vulnerability information in images for evaluating urban building flood vulnerability\cite{47}. This integration of different perspective information provides a more comprehensive and detailed view for safety analysis, significantly improving the overall assessment capability of scene safety and ensuring that the evaluation results more accurately reflect the actual situation of the scene.

Therefore, this paper also adopts multimodal methods, combining the elderly's first-person perspective image information with location information to more accurately identify potential dangerous states for the elderly.

\subsection{anti-loss smart devices and early warning systems}\label{devices}
Current anti-loss smart devices and early warning systems primarily collect location information of the elderly through portable devices and use various technologies to analyze and determine potential dangers, issuing alerts accordingly. These systems include real-time positioning systems using STM32 controllers and SIM868 positioning modules\cite{12}, smart clothing integrating GPS and Beidou Navigation Systems\cite{19}, Bluetooth-based anti-loss alarms\cite{16}, self-powered anti-loss devices combining dual distance measurement and Beidou positioning\cite{14}, Safe Sentry positioning necklaces for caregivers\cite{15}, mobile robot tracking systems\cite{13}, location solutions combining battery-free wearable tags and passive RFID technology\cite{17}, and GEM geofencing systems relying on ambient RF signals\cite{18}. Although these solutions span various fields, including IoT, mobile application development, wearable technology, and robotics, they generally focus on the warning function and often fail to provide caregivers with an accurate display of the elderly's location or surrounding environment, lacking visual monitoring.

To address this limitation, this study proposes an innovative solution that displays the elderly's first-person perspective images and location information in real-time on the main interface of a mobile application, achieving visual monitoring. Additionally, our solution introduces a multi-level proactive warning mechanism based on the elderly's dangerous state, enabling a more timely understanding of their safety status.
\begin{figure*}[t]
\centering
\includegraphics[width=\linewidth]{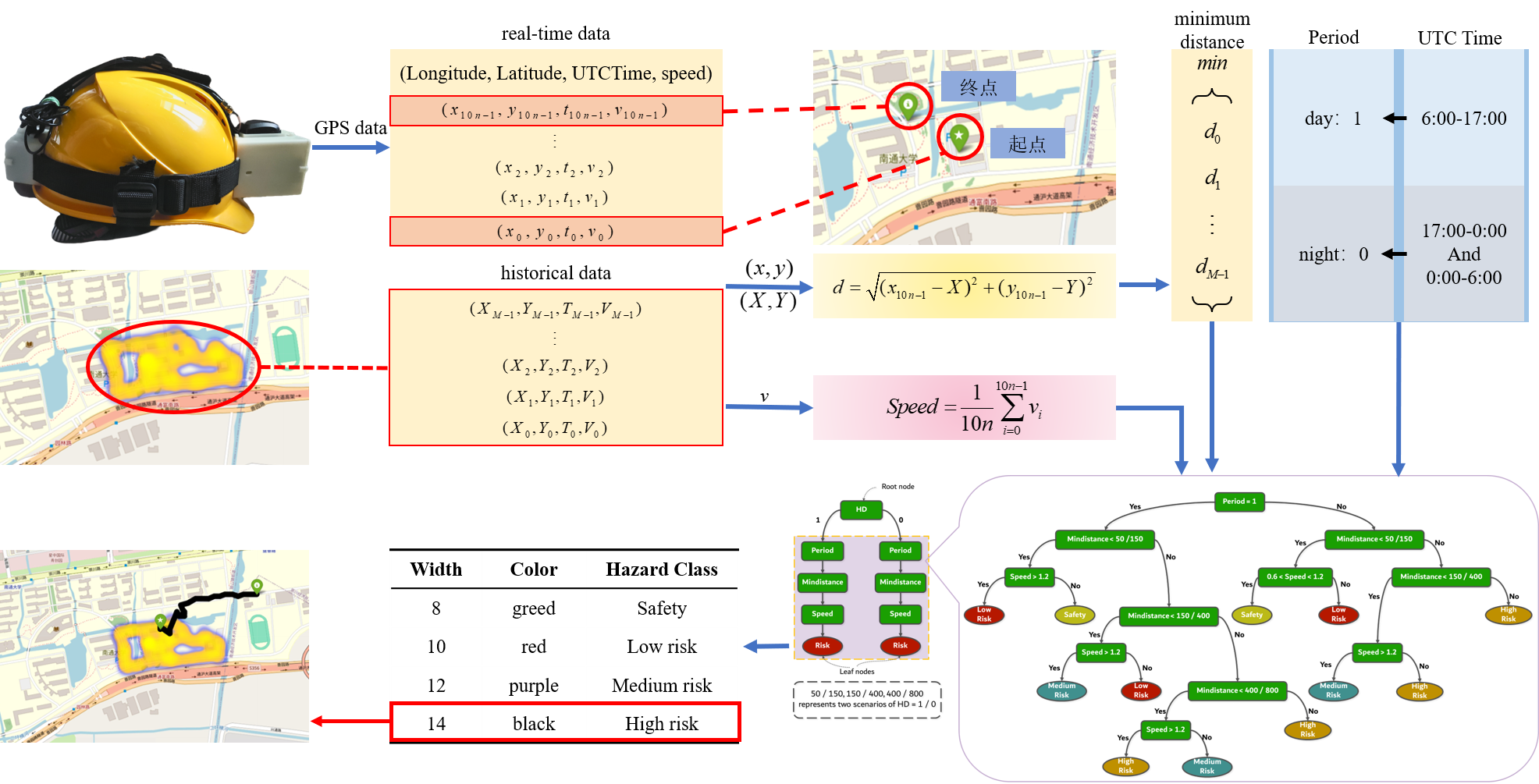}
\caption{The structure diagram of the scene-location multimodal dangerous state recognition network.}
\label{fig3}
\end{figure*}

\section{scene-location multimodal dangerous state recognition network}
In this paper, to more comprehensively evaluate the current dangerous state of the elderly, we propose a scene-location multimodal dangerous state recognition network based on image and location information. As shown in Figure \ref{fig2}, our proposed model includes four key modules: the Map Generation Module (MGM), the Unimodal Feature Extraction Module (UFE), the Multimodal Feature Fusion Module (MFF), and the Hazardous State Recognition Module (HSR). First, the MGM module converts location information into more intuitive map images. Next, the map images and scene images are processed in parallel in the UFE module to extract features. Then, the obtained features are sent to the MFF module for comprehensive multimodal feature fusion. Finally, the HSR module analyzes the fused features to identify the current dangerous state. The following sections will provide a more detailed description and discussion of the MGM, UFE, MFF, and HSR modules.
\subsection{pseudo map generation module}
Select valuable information from the collected location data: latitude, longitude, UTC time, and ground speed \( (k) \) in knots. Determine the elderly's movement trajectory through changes in latitude and longitude, divide the time of day into daytime and nighttime using the UTC time field, and convert the ground speed into the elderly's travel speed \( (v) \) in meters per second \( (m/s) \) using the following conversion formula:

\[ v = k \times \frac{1852}{3600} \]

Where \( k \) is the ground speed in knots, 1852 is the conversion factor from nautical miles to meters, and 3600 is the conversion factor from hours to seconds.

The specific process is shown in Figure \ref{fig3}. Using Python's Folium library, an original map is generated that displays the elderly's location changes every 5 minutes. The GPS module of the smart helmet collects location information every 30 seconds, so the number of data points on each map is a multiple of 10. The first position collected by the smart helmet for each outing \((x_0, y_0)\) is used as the starting point, with its UTC time recorded as \(t_0\), where \(x\) and \(y\) represent the latitude and longitude of the elderly’s location. The ending time for the elderly’s position displayed on each map is \(t_0 + 300n\), where \(n\) is the nth map generated during this outing. The endpoint coordinates are represented as \((x_{10n-1}, y_{10n-1})\). 

The trajectory from \((x_0, y_0)\) to \((x_{10n-1}, y_{10n-1})\) is represented as the vector \(\overrightarrow{p_0}\), and the elderly’s movement trajectory on each map can be represented as the set \(\{\overrightarrow{p_0}, \ldots, \overrightarrow{p_{10n-1}}\}\). The starting position \((x_0, y_0)\) is marked with a star, and the endpoint position \((x_{10n-1}, y_{10n-1})\) is marked with an "i".

The color and thickness of the trajectory lines are determined based on the elderly’s travel speed, the time of day they are traveling, and the distance from familiar areas, to identify the danger level in the elderly’s current location information.
\subsubsection{The average speed of the elderly during this period (Speed)}
 As shown in Table \ref{table1}, the estimated walking speed for elderly individuals over 65 years old is between 0.6-1.2 m/s. The normal average speed for elderly individuals is defined as \( 0.6 m/s < Speed < 1.2 m/s \). \( Speed > 1.2 m/s \) indicates a fast speed, and  \( Speed < 0.6 m/s \) indicates a slow speed\cite{48}. Whether the speed is too fast or too slow, it can result in a higher danger level for the elderly compared to when they walk at a normal speed.
 
\begin{table}[ht]
\caption{Average Walking Speed of Elderly in Different States}
\label{table1}
\centering
\begin{tabular}{cc}
   \toprule
   Speed  & State  \\
   \midrule
   Less than 0.6 m/s  & Slow \\
   Between 0.6-1.2 m/s  & Normal \\
   Greater than 1.2 m/s & Fast \\
   \bottomrule
\end{tabular}
\end{table}

\subsubsection{The time period when the elderly go out (Period)}
 either daytime or nighttime. Daytime: Period = 1, Nighttime: Period = 0.
\subsubsection{The minimum distance}
If the system has recent historical data of the elderly's outings (HD = 1), the historical locations' latitude and longitude coordinates are denoted as \((x_m, y_m)\), where \(m = 0, \ldots, M-1\), with a total of \(M\) historical data points. Use equation (2) to calculate the distance from each map's endpoint \((x_n, y_n)\) to the historical locations \((x_m, y_m)\), \(d_0, d_1, \ldots, d_M\).

\begin{equation}
 d_i = \sqrt{(x_n - x_m)^2 + (y_n - y_m)^2}   
\end{equation}

Find the minimum distance \( \text{MinDistance} \). \(\text{MinDistance}\) is divided into four levels: \(\text{MinDistance} < 50 \, m\), \(\text{MinDistance} < 150 \, m\), \(\text{MinDistance} < 400 \, m\), \(\text{MinDistance} > 400 \, m\).

If there is no recent historical data of the elderly's outings (HD = 0), calculate the distances as follows: 

\[
\text{MinDistance} = \sqrt{(x_n - x_0)^2 + (y_n - y_0)^2}
\]

where \( x_0, y_0 \) are the starting coordinates of the outing.
The distance from each map's endpoint to the starting point is divided into four levels: \(\text{MinDistance} < 50 \, m\), \(\text{MinDistance} < 400 \, m\), \(\text{MinDistance} < 800 \, m\), and \(\text{MinDistance} > 800 \, m\), as shown in Table \ref{table2}. 

\begin{table}[ht]
\caption{Thresholds for MinDistance in Two Scenarios}
\label{table2}
\centering
\begin{tabular}{cc}
   \toprule
   HD = 0 & HD = 1  \\
   \midrule
   \(< 150\)  & \(< 50\) \\
   \(< 400\)  & \(< 150\) \\
   \(< 800\)  & \(< 400\) \\
   \(> 800\)  & \(> 400\) \\
   \bottomrule
\end{tabular}
\end{table}

A danger level decision tree is constructed based on the three indicators: Speed, Period, and MinDistance, to classify the danger level of the location information, as shown in Table \ref{table3}.

\begin{table}[ht]
\caption{Track Parameters for Different Danger Levels}
\label{table3}
\centering
\begin{tabular}{ccc}
   \toprule
   Danger Level & Track Color & Line Width \\
   \midrule
   Safe & Green & 8 \\
   Low & Red & 10 \\
   Medium & Purple & 12 \\
   \bottomrule
\end{tabular}
\end{table}

When the danger level of the location information is Safe, the travel track color is green, and the track line width is 8.

When the danger level is Low, the travel track color is red, and the track line width is 10.

When the danger level is Medium, the travel track color is purple, and the track line width is 12.

When the danger level is High, the travel track color is black, and the track line width is 14.

\begin{figure}[ht]
\centering
\includegraphics[width=\linewidth]{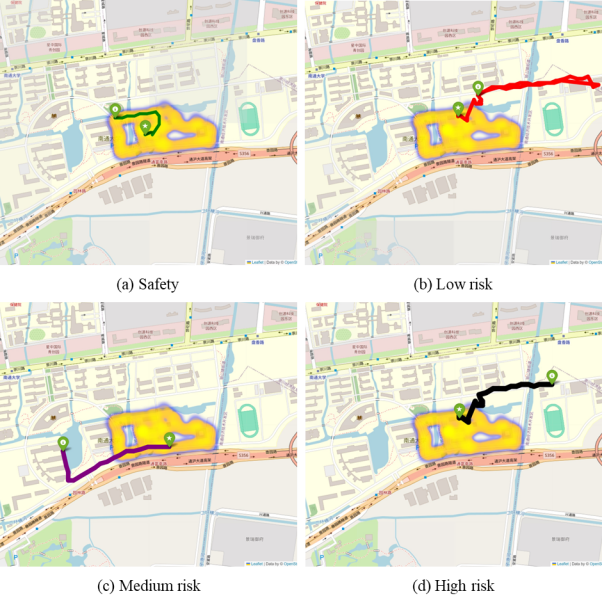}
\caption{Sample Map Generated by the Pseudo Map Generator}
\label{fig4}
\end{figure}

If HD = 1, the historical outing locations are marked on the map as heat points. The resulting heat map area represents familiar regions, and the rest of the area represents unfamiliar regions. A sample generated map is shown in Figure \ref{fig4}.

\subsection{unimodal feature extraction module (UFE)}

The UFE module is divided into the Scene Image Feature Extraction Module (SUFE) (the orange branch in Figure \ref{fig2}) and the Map Image Feature Extraction Module (MUFE) (the blue branch in Figure \ref{fig2}). A pretrained ResNet\cite{49} is used as SUFE to extract scene features. A self-attention and LSTM hybrid model is cascaded to form MUFE to extract map features. The Self-Attention Network (SAN)\cite{50} uses similar block connections and residual connections as ResNet to build the network. The SAN block unit starts with an input map image data of shape \( (3, 227, 227) \). The SAN is composed of several stacked SAN blocks.

\subsubsection{Self-Attention Module (SAN)}
Taking the middle SAN block as an example, given an input feature \( x \) of shape \( (C', H', W') \), we can capture more complex and abstract features \( F_{out} \) through the following steps:

First, apply three separate 1x1 convolutions \(\varphi\), \(\psi\), and \(\beta\) to the feature \( x \). The convolutions \(\varphi\) and \(\psi\) generate two features \(\varphi(x)\) and \(\psi(x)\) of shape \( (C', H, W) \). The convolution \(\beta\) generates a feature \(\beta(x)\) of shape \( (C'', H, W) \), where \( C'' = \frac{C}{4} \).

Next, perform position encoding for each pixel in the feature map, denoted with subscripts to represent the pixel's position, such as \(\varphi(x_i)\) and \(\psi(x_j)\). Compute the correlation between pixel \(i\) and the pixels in its neighboring region \(R(i)\). Concatenate \(\varphi(x_i)\) and \(\psi(x_j)\), \(j \in R(i)\) to obtain the feature vector \(\delta(x_i, x_j)\), which is represented by equation \eqref{delta(x_i, x_j)}:
% % 整个公式只有一个编号
\begin{equation}\label{delta(x_i, x_j)}
  \delta(x_i, x_j) = \text{concat}(\varphi(x_i), \psi(x_j))
\end{equation}
Here, the concatenated feature vector \(\delta(x_i, x_j)\) captures the relationship between pixel \(i\) and its neighboring pixels within the region \(R(i)\).

Next,  the \(\gamma\) function performs a \{Conv-ReLU-Conv\} operation to map \(\delta(x_i, x_j)\) to the same feature space as \(\beta(x)\). The \(\gamma\) function changes the channel number of \(\delta(x_i, x_j)\) from \( \frac{C}{16} \) to \( \frac{C}{4} \), obtaining the relationship weight \( F_j \) between \(\varphi(x_i)\) and \(\psi(x_j)\) at any position within its neighborhood \(R(i)\), as shown in equation \eqref{F_j}:
\begin{equation}\label{F_j}
  F_j = \gamma(\delta(x_i, x_j))
\end{equation}
This process captures the relationship weights \(F_j\) between \(\varphi(x_i)\) and the \(\psi(x_j)\) of its neighboring pixels within the region \(R(i)\).

Next, perform the Hadamard product between \(F_j\) and \(\beta(x_j)\) and sum the results to obtain the attention weight feature vector \(Sa(x)\) of shape \(H \times W \times \frac{C}{4}\), as shown in equation \eqref{Sa(x)}:
\begin{equation}\label{Sa(x)}
  Sa(x) = \sum_{j \in R(i)} F_j \odot \beta(x_j) \
\end{equation}

Finally, pass \(Sa(x)\) through Batch Normalization (BN), ReLU, and a convolutional layer to reshape it into \(H \times W \times C\), resulting in \(F(x)\). Add \(F(x)\) to the input feature \(x\), and then pass the result through BN and ReLU layers to obtain the output feature \(F_{out}\), as shown in equations \eqref{F(x)} and \eqref{F_{out}}:
\begin{equation}\label{F(x)}F(x)=Conv(Relu(BN(Sa(x)))\end{equation}
\begin{equation}\label{F_{out}}F_{out} = \text{ReLU}(\text{BN}(F(x) + x))\end{equation}
\subsubsection{Long Short Term Memory (LSTM)}The features output by the SAN are reshaped into a format of (seq\_len, batch, input\_size) and input into the LSTM to capture temporal features between map sequences. We construct a 3-layer LSTM model with seq\_len = 2. The internal mechanisms of the LSTM can be represented by equations \eqref{c^{t}}, \eqref{h^{t}}, and \eqref{y^{t}}:
\begin{equation}\label{c^{t}}c^{t} = z^{f} \odot c^{t-1} + z^{i} \odot z\end{equation}
\begin{equation}\label{h^{t}}h^{t} = z^{o} \odot \tanh(c^{t})\end{equation}
\begin{equation}\label{y^{t}}y^{t} = \sigma(W^{'}h^{t})\end{equation}

Here:
\( c^{t} \) is the cell state at time step \( t \).
\( z^{f} \), \( z^{i} \), and \( z^{o} \) are the forget gate, input gate, and output gate activations, respectively.
\( h^{t} \) is the hidden state at time step \( t \).
\( \sigma \) is the sigmoid activation function.
\( \odot \) denotes element-wise multiplication.
\( \tanh \) is the hyperbolic tangent function.
\( W^{'} \) represents the weight matrix applied to the hidden state.
\subsection{Multimodal Feature Fusion Module (MFF)}

First, the output features \( F_{1} \) from the SAN and \( F_{2} \) from the LSTM are concatenated along the channel dimension to obtain the multimodal feature \( F_{12} \), as shown in equation \eqref{F_{12}}:
\begin{equation}\label{F_{12}}F_{12} = \text{concat}(F_{1}, F_{2})\end{equation}

This concatenation combines the spatial features extracted by the SAN with the temporal features captured by the LSTM, resulting in a comprehensive feature representation that leverages both spatial and temporal information.

Next, concatenate the global scene features \( F_{3} \) output by the SUFE module with \( F_{12} \) along the channel dimension to fuse the information from both modalities, resulting in \( F_{multi} \). The calculation can be represented as equation \eqref{F_{multi}}:

\begin{equation}\label{F_{multi}}F_{multi} = \text{concat}(F_{3}, F_{12})\end{equation}

The fused features \( F_{multi} \) are then input into the SENet channel attention module. This module determines the importance (i.e. weights) of the input features across different channels through network operations, focusing on channels that contain key information while giving less attention to channels with less information. This process effectively enhances the feature representation capability.

To facilitate subsequent dangerous state recognition, the features output by the SENet module are flattened into a column vector, as represented by equation \eqref{F_c}:
\begin{equation}\label{F_c}F_c = \text{Flatten}(\text{SE}(F_{multi}))\end{equation}

This step prepares the features for the final classification or recognition tasks by converting the multi-dimensional feature map into a one-dimensional vector.
% % 整个公式只有一个编号
\subsection{Hazardous State Recognition Module (HSR)}
The Hazardous State Recognition Module performs a sequence of operations: \{Linear - ReLU - Linear - Softmax\}. This module uses two stacked linear layers to map the learned features to the sample label space. We utilize the nn.Sequential module to connect the two linear layers and insert a ReLU activation function after the first linear layer, which helps the network learn more complex functional relationships. The output dimension of the second linear layer is 4, performing another linear transformation and adding a bias term, further converting the features from the previous layer into the probabilities of four hazardous states. The linear layer`s computation can be represented by equation \eqref{y}:
\begin{equation}\label{y}y=x*W^T+b\end{equation}
\section{Anti-Loss Early Warning System Based on Smart Helmet}
This study proposes an innovative intelligent early warning system based on the aforementioned scene-location multimodal dangerous state recognition network. The system integrates a smart helmet equipped with an intelligent camera, a server-side elderly dangerous state recognition network, and a multi-level intelligent early warning software, providing comprehensive monitoring and warning support for the safety of elderly individuals during outings.

When the elderly go out, they wear this smart helmet, which automatically collects first-person perspective image information every 5 minutes and location information every 30 seconds. These data are uploaded to the server via 5G communication technology using FTP and TCP protocols. The image and location information received by the server are then analyzed by the scene-location multimodal dangerous state recognition network to identify the current dangerous state of the elderly. The image information is sent to the Android-based intelligent early warning software via the FTP protocol, while the location information and danger state assessment results are transmitted via the TCP protocol. The software visualizes these information on its interface and triggers corresponding level warnings based on the identified dangerous state.

Subsequent sections will provide detailed introductions to the functions and design of the anti-loss smart helmet and the multi-level intelligent early warning software.

\subsection{Anti-Loss Smart Helmet}
The helmet consists of a safety helmet, an intelligent camera module, and a solar battery. The selected safety helmet features a hard shell and soft lining, designed to prevent accidental head injuries while ensuring wearing comfort for the elderly. The intelligent camera module includes four core components: the camera sensor, GPS module, 5G communication module, and power control module. It uses the G8100 chip as the main controller to manage information collection intervals, specify the server IP for data uploads, and maintain a long TCP connection with the server. The 5G communication module enables real-time transmission of image data and location information to the remote server. The power control module stabilizes the battery voltage or the output voltage from the solar panel. The inclusion of solar batteries provides an eco-friendly and continuous energy solution for the helmet.

\begin{figure}[ht]
\centering
\includegraphics[width=\linewidth]{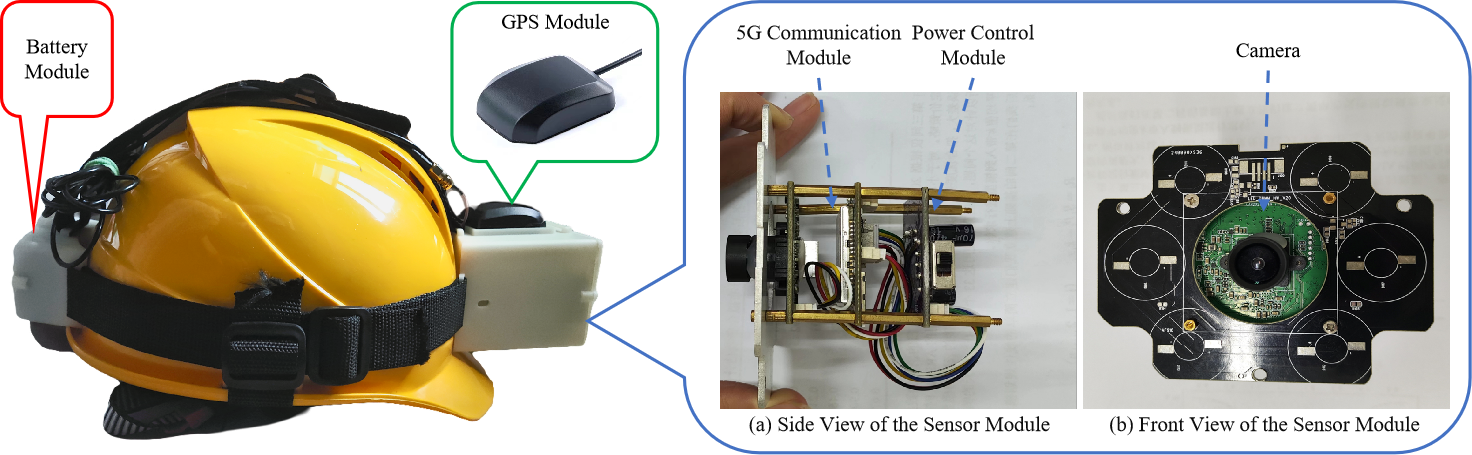}
\caption{Assembly Effect of the Smart Helmet and Specific Details of the Intelligent Camera Module}
\label{fig5}
\end{figure}

The installation process of the smart helmet is as follows: first, place the intelligent camera module, except for the GPS, into the sensor mounting box, and fix the GPS module in the groove on the surface of the mounting box. Then, place the solar battery into the battery mounting box. Install the sensor mounting box on the helmet`s brim and fix the battery mounting box at the back of the helmet. Use fixing straps through the perforations in the helmet body to secure these mounting boxes at the front and rear positions of the helmet. The assembly effect of the smart helmet and the specific details of the intelligent camera module are shown in Figure \ref{fig5}.

\begin{figure}[ht]
\centering
\includegraphics[width=\linewidth]{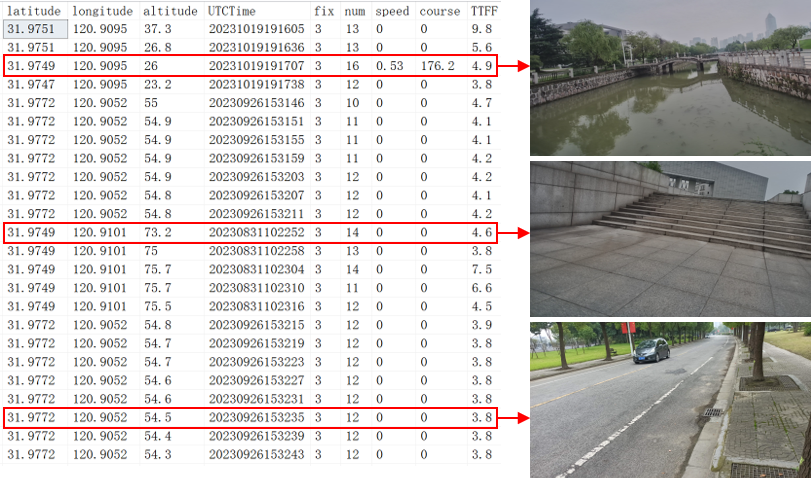}
\caption{Example of Information Collected by the Smart Helmet}
\label{fig6}
\end{figure}

The smart helmet primarily collects information through the camera sensor and GPS module. The camera sensor is responsible for capturing first-person perspective image data from the elderly user and transmitting it to the server via 5G communication technology and FTP protocol for storage. The GPS module collects location information, including latitude, longitude, altitude, UTC time, 2D/3D positioning status, the number of satellites used, ground speed, ground heading, and first positioning time. This information is sent to the server through a TCP long connection and stored in a Microsoft SQL Server database. Examples of the collected image and location information are shown in Figure \ref{fig6}.
\subsection{Mobile Early Warning Application}
The main interface of the mobile early warning application is divided into two sections: (1) Top Section: Real-Time First-Person Perspective Images: Displays live images from the elderly individual`s point of view captured by the smart helmet.
Textual Danger Status: Shows the current danger status of the elderly in text form. (2) Bottom Section: Dynamic Map: Tracks the exact real-time location of the elderly. And displays the activity trajectory within a certain time range using a heat map.
Additionally, the application integrates an advanced proactive warning mechanism. This mechanism automatically executes warning actions based on the analysis results of a specially designed dangerous state recognition network:
\subsubsection{High Danger}
The application directly calls the caregiver.
\subsubsection{Medium Danger}
The application triggers an unmissable pop-up notification to alert the caregiver.
\subsubsection{Low Danger}
The application sends an SMS notification to inform the caregiver.

\begin{figure}[ht]
\centering
\includegraphics[width=\linewidth]{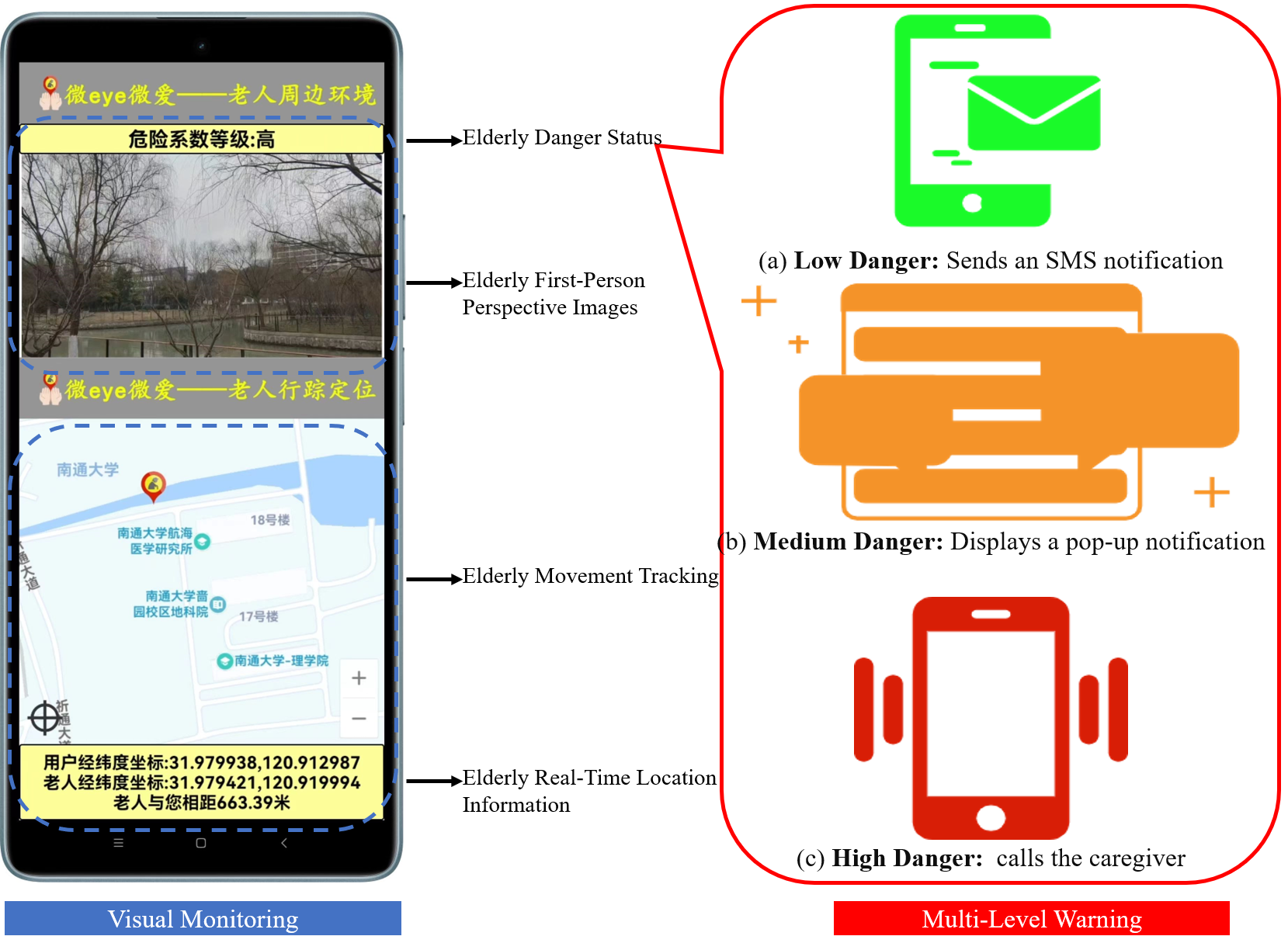}
\caption{An example of the mobile application interface}
\label{fig7}
\end{figure}

The unique aspect of this system is that it not only provides precise location tracking of the elderly but also adds the ability to display first-person perspective images and record their movement trajectory. This offers caregivers a more intuitive and comprehensive monitoring approach, allowing them to understand the elderly individual`s immediate situation and safety needs in greater detail. An example of the mobile application interface and multi-level warning methods is shown in Figure \ref{fig7}.

\section{Experimental Results and Analysis}
\subsection{Experimental Data and Environment}\label{Environment}
In this study, 2930 data sets were collected by having team members wear the smart helmet to simulate elderly outings. The data sets include scene images, map images, and danger status labels. The scene classification covers five common scenes for the elderly: road, stairs, riverside (with railings), woods, and street. The danger level classification results are as follows: Safe (486 images), Low Danger (896 images), Medium Danger (773 images), and High Danger (1075 images). For the map images, the classifications are: Safe (1612 images), Low Danger (1017 images), Medium Danger (499 images), and High Danger (96 images). The comprehensive evaluation of the elderly`s danger status distribution is: Safe (880 sets), Low Danger (888 sets), Medium Danger (363 sets), and High Danger (1093 sets), showing an imbalance in the data distribution. Considering this, the evaluation metrics selected are weighted metrics: \( P_{weight} \), \( R_{weight} \), \( F1_{weight} \). The data is split into training and testing sets in a 9:1 ratio.

The experimental environment was configured with a Linux system, using Python 3.10.9 and Pytorch 2.1.2+cu121. The hardware included an Intel Core i9-10900X CPU, a GeForce RTX 2080ti GPU, and 251GB of memory. 

Data preprocessing involved resizing images to a 227x227 resolution, applying random horizontal flips, and normalizing the training set. The predefined RGB channel mean [0.4725, 0.4652, 0.4438] and standard deviation [0.2471, 0.2447, 0.2542] were used for normalization. Similar processing was applied to the test set.

Feature extraction utilized pretrained ResNet weights. Model training was performed using the Adadelta optimizer with an initial learning rate of 0.01 and a weight decay rate of 0.5. The batch size was set to 10, and the training was conducted for 150 epochs, with the learning rate adjusted every 20 epochs. CrossEntropyLoss() was used as the loss function.

\subsection{Evaluation Metrics}
In this study, we used Precision (P), Recall (R), F1 Score (F1), and Accuracy (Acc) to evaluate the performance of our model. The calculation formulas are as follows:

\begin{equation}\label{P_i}P_i = \frac{TP_i}{TP_i + FP_i}\end{equation}
\begin{equation}\label{R_i}R_i = \frac{TP_i}{TP_i + FN_i}\end{equation}
\begin{equation}\label{F1_i}F1_i = \frac{2 \times P_i \times R_i}{P_i + R_i}\end{equation}

Considering the issue of class imbalance, we used weighted metrics \(P_{weight}\), \(R_{weight}\), and \(F1_{weight}\) to evaluate our model. The calculation formulas are:
\begin{equation}\label{w_i}w_i = \frac{N_i}{L}\end{equation}

where \(N_i\) represents the total number of samples of class \(i\) in the entire dataset, and \(w_i\) represents the proportion of each sample in the entire dataset. \(L\) represents the length of the entire dataset.
\begin{equation}\label{P_{weight}}P_{weight} = \frac{\sum_{i=1}^L P_i \times w_i}{L}\end{equation}

Similarly, \(R_{weight}\) and \(F1_{weight}\) can be calculated.

\subsection{Ablation Experiments}
To verify the effectiveness of our proposed network structure, we designed 10 groups of experiments to explore the following aspects: Map: Performs danger state detection using single-modal map data. S: Performs danger state detection using single-modal scene data. SM: Combines scene and map image features using concatenation for danger state detection. 
SM\_SE: Based on references\cite{51,52,53}, introduces a channel attention mechanism on top of feature concatenation for further danger state detection. 
SM\_LSTM: Referring to references\cite{54,55}, adopts a hybrid CNN-LSTM model framework, replacing the CNN part with ResNet18 for map feature extraction, and combines it with scene image features for danger state recognition. 
SM\_LSTM\_M: According to reference\cite{56}, uses convolutional autoencoders (CAE) and hybrid CNN-LSTM to extract image and semantic features, respectively. This study replaces the original CAE encoder with ResNet18 and uses the same framework to extract spatial and temporal features from the map, then combines them with scene image features to identify danger states. SM\_LSTM\_M\_SE: Further integrates the SENet module on the SM\_LSTM\_M model. R18\_S14: Analyzes the results and data of the aforementioned models by introducing the R18\_S14 model, which replaces the map feature extraction module in SM\_LSTM\_M\_SE with the self-attention mechanism SAN14 to capture subtle changes in map paths, while continuing to use ResNet18 for scene feature extraction. R50\_S14: To further enhance model performance, develops the R50\_S14 model, where the scene feature extraction module is replaced by ResNet50, and the map feature extraction module is SAN14. R50\_S18: Similar to R50\_S14, but uses SAN18 as the map feature extraction module.

\begin{figure}[ht]
\centering
\includegraphics[width=\linewidth]{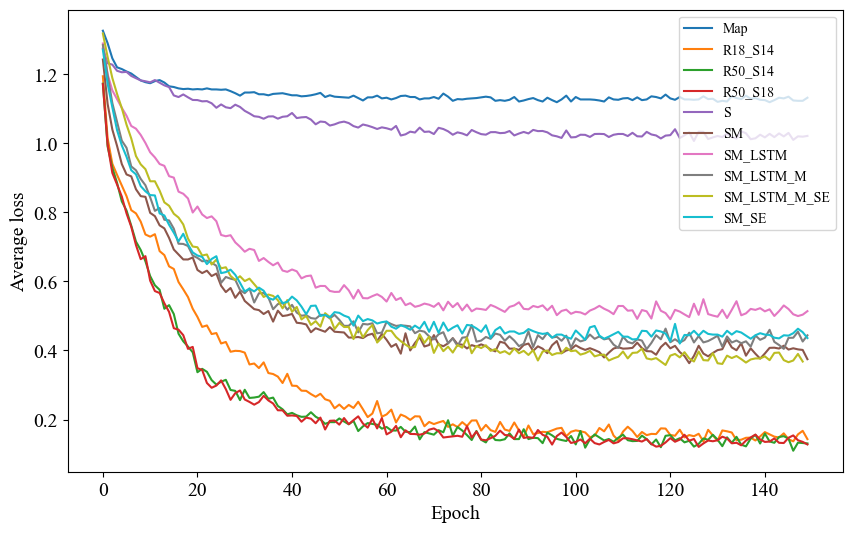}
\caption{Loss Value Changes During Training}
\label{fig8}
\end{figure}

All experiments follow the parameter settings described in Section \ref{Environment}, and the experimental results are detailed in Figure \ref{fig8} and Table \ref{table4}.

\begin{table*}[ht]
\caption{experimental results}
\label{table4}
\centering
\begin{tabular}{cccccc}
   \toprule
    Model & precision & recall & F1-score & Average loss & Accuracy(\%)  \\ 
    \midrule
    \multicolumn{6}{c}{Unimodal} \\
    \midrule
    Map & 0.42 & 0.47 & 0.42 & 1.14  & 47.02   \\ 
    S & 0.52 & 0.54 & 0.52 & 1.64  & 54.30   \\
    \midrule
    \multicolumn{6}{c}{Multimodal}  \\ 
    \midrule
    SM & 0.65 & 0.62 & 0.63 & 0.95  & 62.25   \\ 
    SM\_SE & 0.63 & 0.61 & 0.62 & 0.93  & 60.79   \\ 
    SM\_LSTM & 0.61 & 0.54 & 0.56 & 1.22  & 54.32   \\
    SM\_LSTM\_M & 0.64 & 0.62 & 0.61 & 1.02  & 62.50   \\ 
    SM\_LSTM\_M\_SE & 0.63 & 0.63 & 0.61 & 0.98  & 62.95   \\ 
    \midrule
    \multicolumn{6}{c}{Ours}  \\ 
    \midrule
    R18\_S14 & 0.68 & 0.67 & 0.67 & 1.38  & 67.01   \\ 
    R50\_S14 & 0.69 & 0.69 & 0.69 & 0.83  & 69.05   \\ 
    \pmb{R50\_S18} & \pmb{0.73} & \pmb{0.72} & \pmb{0.72} & \pmb{0.73} & \pmb{72.11}    \\ 
   \bottomrule
\end{tabular}
\end{table*}

As shown in Figure \ref{fig8}, the proposed models utilizing ResNet and self-attention mechanisms for unimodal feature extraction demonstrate superior convergence speed and effectiveness compared to previous models. Analyzing the performance of various models on the test set reveals that multimodal models significantly outperform unimodal models in accurately identifying the danger status of the elderly. Our proposed models exhibit better performance across all evaluation metrics compared to earlier models. Specifically, the R50\_S18 model shows an improvement of 0.1 in precision, 0.09 in recall, 0.11 in F1 score, a reduction of 0.25 in loss value, and an overall accuracy increase of 9.16\% compared to the SM\_LSTM\_M\_SE model.

\subsection{Comparative Experiments with Existing Methods}

One of the innovations in this study is the conversion of text-based location information into map image data for elderly outings. To verify the improvement in the accuracy of dangerous state recognition for the elderly through this innovation, we conducted experiments comparing text and scene-based methods like MDFNet\cite{51}, where the text data is derived from further mining the collected location information, as shown in the results. We conducted comprehensive comparative experiments with other existing multimodal early warning models, as illustrated in the results.

The MDFNet\cite{51} model attempts to perform early warning by fusing text and image data features, but the significant difference in feature dimensions between these two types of data makes it challenging to balance their weights, thereby affecting the overall performance of the model. The FGT-Net\cite{57} model relies on typical sample images to guide the extraction of image features and combines the corresponding text data for classification. This method is suitable for datasets with small inter-group differences but does not match the attributes of our dataset, leading to suboptimal results. The YOLOv5s-DMF\cite{46} model uses YOLOv5s to extract and fuse features from RGB and infrared images. Although it captures the overall features of the scene, it is insufficient to detect subtle changes in the map images, resulting in poor performance on our dataset. The FSA-UNet\cite{47} model extracts features from remote sensing and street view images separately and uses a self-attention mechanism for multimodal fusion. While it also focuses on the overall features of images during the feature extraction phase, the self-attention mechanism significantly enhances accuracy and precision, outperforming the YOLOv5s-DMF model.

Our proposed model uses a self-attention module to capture subtle changes in the elderly`s movement trajectory on the map and employs the ResNet model to extract overall features from the elderly`s first-person perspective images. By fusing these features, our model accurately identifies the dangerous states of the elderly. Experimental results show that compared to existing technologies, our model excels in terms of precision, accuracy, and loss value, making it more suitable for our dataset. This outcome not only demonstrates the superiority of our model but also provides a new technological pathway for the effective identification of dangerous states in the elderly.
\begin{table*}[ht]
\caption{Comparative Experimental Results}
\label{table5}
\centering
\begin{tabular}{cccccc}
   \toprule
    Model & precision & recall & F1-score & Average loss & Accuracy(\%)  \\ 
    \midrule
        MDFNet\cite{51}  & 0.51  & 0.56  & 0.53  & 1.12  & 55.89   \\ 
        FGT-Net\cite{57}  & 0.62  & 0.62  & 0.61  & 1.10  & 62.12   \\ 
        YOLOv5s-DMF\cite{46}  & 0.58 & 0.57 & 0.58 & 1.21 & 61.10  \\ 
        FSA-UNet\cite{47}   & 0.65 & 0.66 & 0.64 & 1.09 & 65.12  \\ 
    \pmb{R50\_S18(ours)} & \pmb{0.73} & \pmb{0.72} & \pmb{0.72} & \pmb{0.73} & \pmb{72.11}    \\ 
   \bottomrule
\end{tabular}
\end{table*}

\subsection{Real-World System Testing Experiment}

\begin{figure}[ht]
\centering
\includegraphics[width=\linewidth]{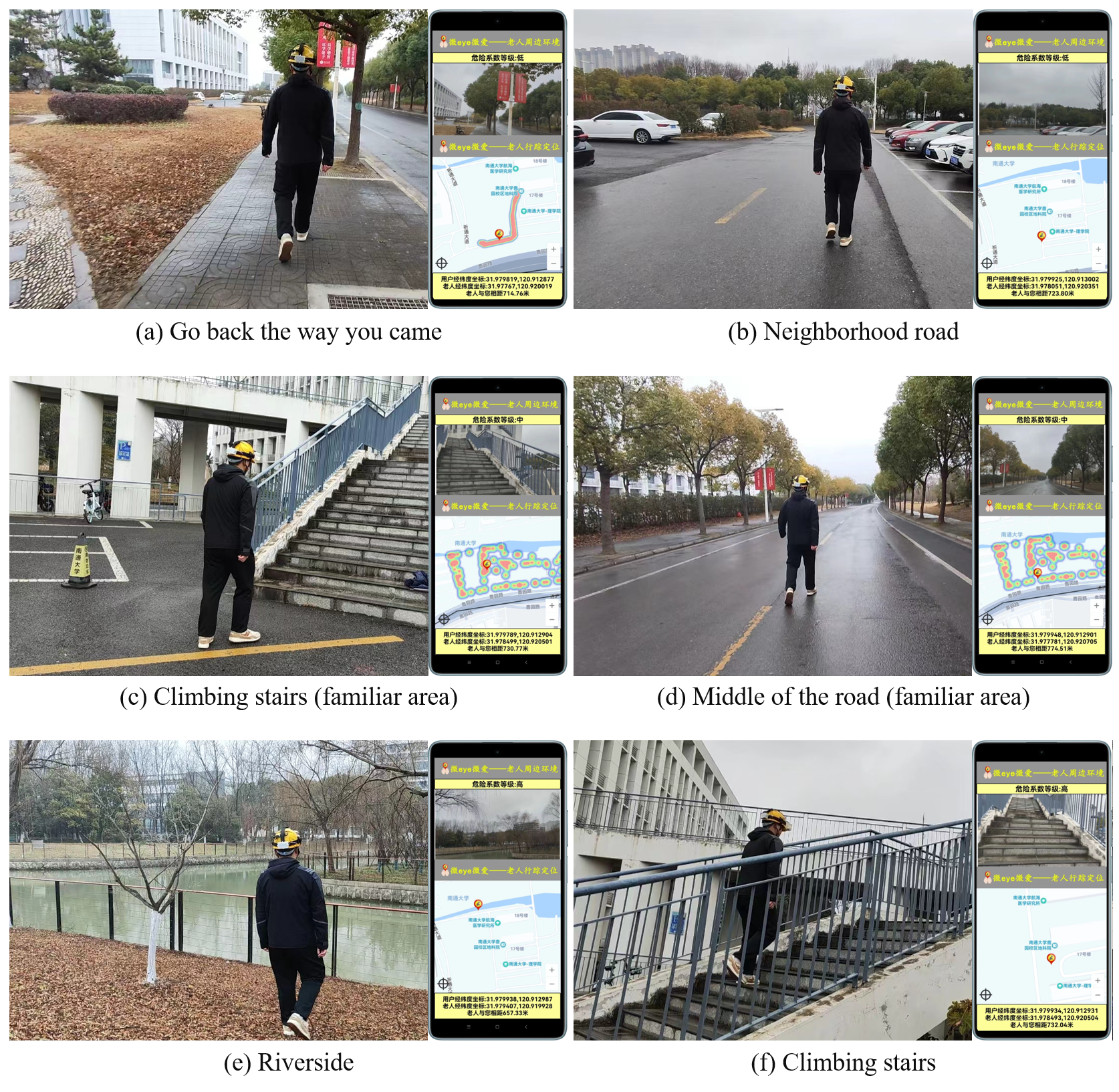}
\caption{Example of Mobile Application Interface from Simulation Experiments}
\label{fig9}
\end{figure}

Our team conducted real-world testing using the developed device, as shown in Figure \ref{fig9}. The different subfigures in Figure \ref{fig9} depict various scenarios: (a) Describes an elderly person returning home. Although the distance from home is significant, the elderly individual is returning along the same route, resulting in a low danger status. (b) Shows an elderly person wearing the helmet for the first time while crossing the road in their residential area. Despite being close to home, the first-time use of the helmet results in a low danger status. (c) Describes an elderly person going up the stairs in a familiar environment. Considering the high risk associated with elderly individuals using stairs, the danger status is classified as medium. (d) Simulates an elderly person walking in the center of a familiar road. Although the location is familiar, being in the middle of the road results in a medium danger status. (e) Simulates an elderly person walking towards the riverside. In this situation, the danger status is classified as high. (f) Simulates an elderly person using the helmet for the first time while going up the stairs. This situation also results in a high danger status.

Compared to existing systems (Section \ref{devices}), our system adds map-scene visualization functionality and a multi-level warning mechanism. However, in terms of processing time, the difference from existing systems is negligible. Therefore, our system still holds significant advantages and application prospects. The detailed results and comparative performance are shown in Table \ref{table6}.
\begin{table*}[ht]
\caption{Performance Comparison with Existing Systems}
\label{table6}
\centering
\begin{tabular}{ccccc}
   \toprule
        System & Location Tracking & Map-Scene Visualization & Multi-Level Warning Mechanism & Processing Time (s) \\ 
    \midrule
        Wang\cite{12}   & \Checkmark & \XSolid & \Checkmark & 5  \\ 
        Zhuo\cite{18}   & \Checkmark & \XSolid & \XSolid & 6 \\ 
        Liang\cite{15}   & \Checkmark & \XSolid & \XSolid & 4  \\ 
        Ours  & \Checkmark & \Checkmark & \Checkmark & 6 \\    
    \bottomrule
\end{tabular}
\end{table*}

\section{Conclusion}
This study proposes a multimodal danger state recognition method based on scene images and location information. Comparisons with existing methods demonstrate that our approach excels in precision, accuracy, and loss values. Additionally, our team designed and developed an intelligent danger early warning system based on this method. This system not only accurately locates the elderly but also visually displays their first-person perspective images and location information on a mobile application for caregivers to monitor. When the system detects that the elderly are in different danger states, it automatically activates the corresponding warning mechanisms, enabling caregivers to precisely understand the elderly’s current status.

The development of this method and system significantly increases the freedom of elderly individuals and greatly enhances the efficiency of caregivers. In summary, our proposed multimodal danger state recognition method and its early warning system offer an efficient and practical new strategy for addressing the problem of elderly individuals going missing. This approach holds significant social value and broad application potential.

% or


%\appendices

%\section{}

%\section*{Acknowledgment}


\ifCLASSOPTIONcaptionsoff
  \newpage
\fi



%\bibliographystyle{IEEEtran}

% \bibliographystyle{plain}
\bibliographystyle{unsrt}
\bibliography{1-57}

%\end{IEEEbiography}


%\newpage



%\vfill


\end{document}